\documentclass[11pt,letterpaper]{article}
\usepackage{emnlp2017}
\usepackage{times}
\usepackage{latexsym}
\usepackage{multirow}
\usepackage{graphicx} 
\usepackage{amsmath}

\emnlpfinalcopy

\def\emnlppaperid{***}

\title{SearchQA: A New Q\&A Dataset \\ Augmented with Context from a Search Engine} 

\author{Matt Dunn \\ Center for Data Science, NYU \And 
        Levent Sagun \\ Courant Institute, NYU \And 
        Mike Higgins \\Center for Data Science, NYU 
        \AND
        V. U\u{g}ur G\"uney \\ Senior Data Scientist, Driversiti \And
        Volkan Cirik \\ School of Computer Science, CMU 
        \AND
        Kyunghyun Cho \\ Courant Institute and Center for Data Science, NYU}

\date{}

\begin{document}

\maketitle

\begin{abstract}
We publicly release a new large-scale dataset, called SearchQA, for machine comprehension, or question-answering. Unlike recently released datasets, such as DeepMind CNN/DailyMail and SQuAD, the proposed SearchQA was constructed to reflect a full pipeline of general question-answering. That is, we start not from an existing article and generate a question-answer pair, but start from an existing question-answer pair, crawled from J! Archive, and augment it with text snippets retrieved by Google. Following this approach, we built SearchQA, which consists of more than 140k question-answer pairs with each pair having 49.6 snippets on average. Each question-answer-context tuple of the SearchQA comes with additional meta-data such as the snippet's URL, which we believe will be valuable resources for future research. We conduct human evaluation as well as test two baseline methods, one simple word selection and the other deep learning based, on the SearchQA. We show that there is a meaningful gap between the human and machine performances. This suggests that the proposed dataset could well serve as a benchmark for question-answering.
\end{abstract}

\section{Introduction}
\label{sec:intro}

One of the driving forces behind the recent success of deep learning in challenging tasks, such as object recognition~\citep{krizhevsky2012imagenet}, speech recognition~\citep{xiong2016achieving} and machine translation~\citep{bahdanau2014neural}, has been the increasing availability of large-scale annotated data.

This observation has also led to the interest in building a large-scale annotated dataset for question-answering. In 2015, \citet{bordes2015large} released a large-scale dataset of 100k open-world question-answer pairs constructed from Freebase, and \citet{hermann2015teaching} released two datasets, each consisting of closed-world question-answer pairs automatically generated from news articles. The latter was followed by \citet{hill2015goldilocks}, \citet{rajpurkar2016squad} and \citet{onishi2016did}, each of which has released a set of large-scale closed-world question-answer pairs focused on a specific aspect of question-answering.

Let us first take a step back, and ask what a full end-to-end pipeline for question-answering would look like. A general question-answering system would be able to answer a question about any domain, based on the world knowledge. This system would consist of three stages. A given question is read and reformulated in the first stage, followed by information retrieval via a search engine. An answer is then synthesized based on the query and a set of retrieved documents.

We notice a gap between the existing closed-world question-answering data sets and this conceptual picture of a general question-answering system. The general question-answering system must deal with a noisy set of retrieved documents, which likely consist of many irrelevant documents as well as semantically and syntactically ill-formed documents. On the other hand, most of the existing closed-world question-answering datasets were constructed in a way that the context provided for each question is guaranteed relevant and well-written. This guarantee comes from the fact that each question-answer-context tuple was generated starting from the context from which the question and answer were extracted.

In this paper, we build a new closed-world question-answering dataset that narrows this gap. Unlike most of the existing work, we start by building a set of question-answer pairs from {\it Jeopardy!}. We augment each question-answer pair, which does not have any context attached to it, by querying Google with the question. This process enables us to retrieve a realistic set of relevant/irrelevant documents, or more specifically their snippets. We filter out those questions whose answers could not be found within the retrieved snippets and those with less than forty web pages returned by Google. We end up with 140k+ question-answer pairs, and in total 6.9M snippets.\footnote{
The dataset can be found at \url{https://github.com/nyu-dl/SearchQA}.
}

We evaluate this new dataset, to which we refer as SearchQA, with a variant of recently proposed attention sum reader \citep{kadlec2016text} and with human volunteers. The evaluation shows that the proposed SearchQA is a challenging task both for humans and machines but there is still a significant gap between them. This suggests that the new dataset would be a valuable resource for further research and advance our ability to build a better automated question-answering system.

%

\section{SearchQA}

\paragraph{Collection} A major goal of the new dataset is to build and provide to the public a machine comprehension dataset that better reflects a noisy information retrieval system. In order to achieve this goal, we need to introduce a natural, realistic noise to the context of each question-answer pair. We use a production-level search engine --Google-- for this purpose. 

We crawled the entire set of question-answer pairs from {\it J! Archive}\footnote{\url{http://j-archive.com}} which has archived all the question-answer pairs from the popular television show Jeopardy!. We used the question from each pair to query Google in order to retrieve a set of relevant web page snippets. The relevancy in this case was fully determined by an unknown, but in-production, algorithm underlying Google's search engine, making it much closer to a realistic scenario of question-answering. 

\paragraph{Cleaning} Because we do not have any control over the internals of Google search engine, we extensively cleaned up the entire set of question-answer-context tuples. First, we removed any snippet returned that included the air-date of the Jeopardy! episode, the exact copy of the question, or a term ``Jeopardy!'', ``quiz'' or ``trivia'', to ensure that the answer could not be found trivially by a process of word/phrase matching. Furthermore, we manually checked any URL, from which these removed snippets were taken, that occurs more than 50 times and removed any that explicitly contains Jeopardy! question-answer pairs. 

Among the remaining question-answer-context tuples, we removed any tuple whose context did not include the answer. This was done mainly for computational efficiency in building a question-answering system using the proposed dataset. We kept only those tuples whose answers were three or less words long.

\paragraph{Basic Statistics}

After all these processes, we have ended up with 140,461 question-answer pairs. Each pair is coupled with a set of 49.6$\pm$2.10 snippets on average. Each snippet is 37.3$\pm$11.7 tokens long on average. Answers are on average 1.47$\pm$0.58 tokens long. There are 1,257,327 unique tokens. 

\paragraph{Meta-Data}

\begin{figure}[t]
\centering
\includegraphics[width=\columnwidth]{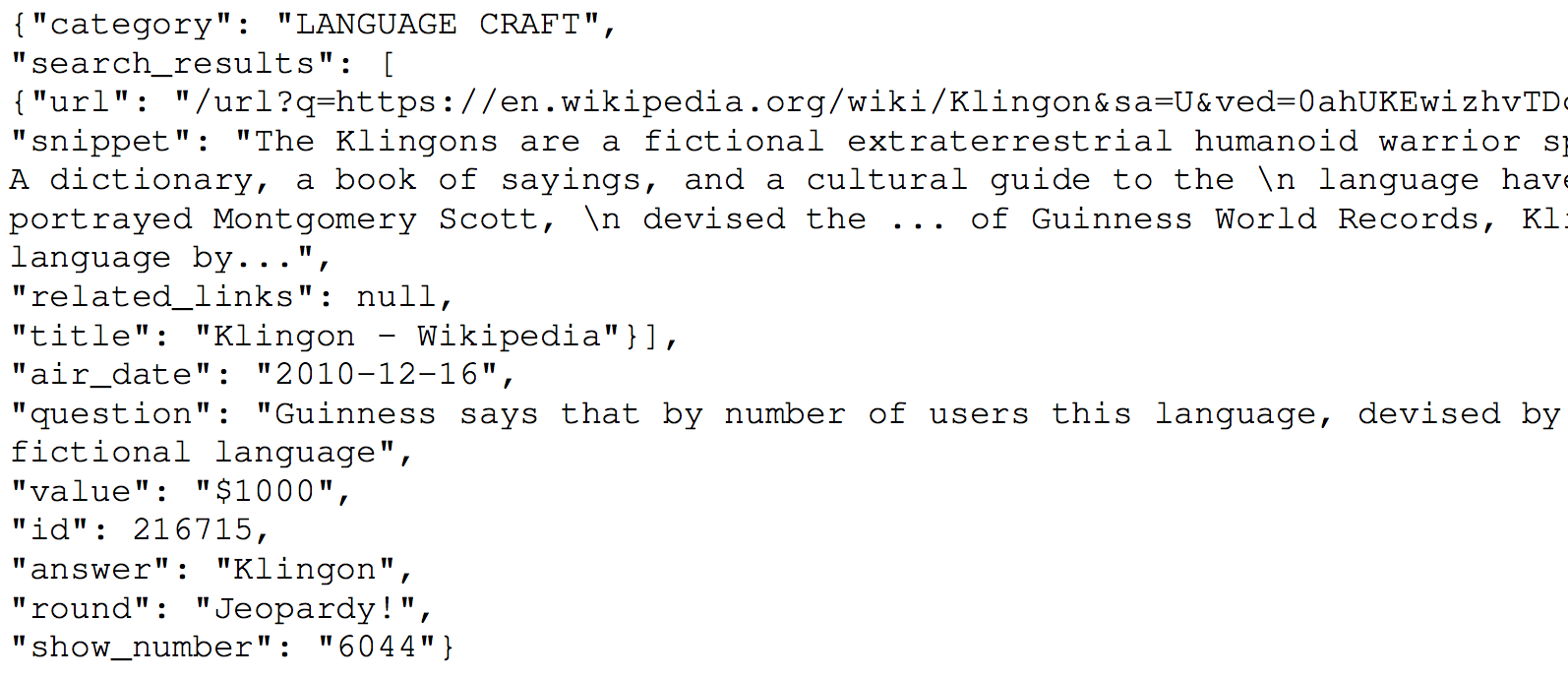}

\vspace{-6mm}
\caption{\label{fig:example} One example in .json format.}

\vskip -6mm
\end{figure}

We collected for each question-answer-context tuple additional metadata from Jeopardy! and returned by Google. More specifically, from Jeopardy! we have the category, dollar value, show number and air date for each question. From Google, we have the URL, title and a set of related links (often none) for each snippet. Although we do not use them in this paper, these items are included in the public release of SearchQA and may be used in the future. An example of one question-answer pair with just one snippet is presented in Fig.~\ref{fig:example}.

\paragraph{Training, Validation and Test Sets}

In order to maximize its reusability and reproducibility, we provide a predefined split of the dataset into training, validation and test sets. One of the most important aspects in question-answering is whether a question-answering machine would generalize to unseen questions {\it from the future}. We thus ensure that these three sets consist of question-answer pairs from non-overlapping years, and that the validation and test question-answer pairs are from years later than the training set's pairs. The training, validation and test sets consist of 99,820, 13,393 and 27,248 examples, respectively. Among these, examples with unigram answers are respectively 55,648, 8,672 and 17,056.


\section{Related Work}


\paragraph{Open-World Question-Answering}

An open-world question-answering dataset consists of a set of question-answer pairs and the knowledge database. It does not come with an explicit link between each question-answer pair and any specific entry in the knowledge database. A representative example of such a dataset is SimpleQA by \citep{bordes2015large}. SimpleQA consists of 100k question-answer pairs, and uses Freebase as a knowledge database. The major limitation of this dataset is that all the questions are {\it simple} in that all of them are in the form of (subject, relationship, ?). 

\paragraph{Closed-World Question-Answering}

Although we use open-world snippets, the final SearchQA is a closed-world question-answering dataset since each question can be answered entirely based on the associated snippets. 
One family of such datasets includes Children's Book dataset~\citep{hill2015goldilocks}, CNN and DailyMail~\citep{hermann2015teaching}. Each question-answer-context tuple in these datasets was constructed by first selecting the context article and then creating a question-answer pair, where the question is a sentence with a missing word and the answer is the missing word. This family differs from SearchQA in two aspects. First, in SearchQA we start from a question-answer pair, and, second, our question is not necessarily of a fill-in-a-word type. 

Another family is an extension of the former family of datasets. This family includes SQuAD~\citep{rajpurkar2016squad} and NEWSQA~\citep{trischler2016newsqa}. Unlike the first family, answers in this family are often multi-word phrases, and they do not necessarily appear as they are in the corresponding context. In contrast, in SearchQA we ensure that all multi-word phrase answers appear in their corresponding context. Answers, often as well as questions, are thus often crowd-sourced in this family of datasets. Nonetheless, each tuple in these datasets was however also constructed starting from a corresponding context article, making them less realistic than the proposed SearchQA. 

MS MARCO~\citep{nguyen2016ms}--the most recently released dataset to our knowledge-- is perhaps most similar to the proposed SearchQA. \citet{nguyen2016ms} selected a subset of actual user-generated queries to Microsoft Bing that correspond to questions. These questions are augmented with a manually selected subset of snippets returned by Bing. The question is then answered by a human. Two major differences between MS MARCO and SearchQA are the choice of questions and search engine. We believe the comparison between MS MARCO and the proposed SearchQA would be valuable for expanding our understanding on how the choice of search engines as well as types of questions impact question-answering systems in the future. 


\section{Experiments and Results}

As a part of our release of SearchQA, we provide a set of baseline performances against which other researchers may compare their future approaches. Unlike most of the previous datasets, SearchQA augments each question-answer pair with a {\it noisy}, {\it real} context retrieved from the largest search engine in the world. This implies that the human performance is not necessarily the upper-bound but we nevertheless provide it as a guideline. 

\begin{table}[t]
\centering
\begin{tabular}{c c c}
Answer & Unigram & $n$-gram \\
\hline\hline
Per-question Average & 66.97\% & 42.86\% \\
Per-user Average & 64.85\% & 43.85\% \\
Per-user Std. Dev. & 8.16\% & 10.43\% \\
F1 score (for $n$-gram) & - & 57.62 \%
\end{tabular}

\vspace{-2mm}
\caption{
\label{tab:human}
The accuracies achieved by the volunteers.
}

\vskip -6mm
\end{table}

\subsection{Human Evaluation}

We designed a web interface that displays a query and retrieved snippets and lets a user select an answer by clicking words on the screen. A user is given up to 40 minutes to answer as many questions as possible. We randomly select question-answer-context pairs from the test set. 

We recruited thirteen volunteers from the master's program in the Center for Data Science at NYU. They were uniform-randomly split into two groups. The first group was presented with questions that have single-word (unigram) answers only, and the other group with questions that have either single-word or multi-word ($n$-gram) answers. On average, each participant answers 47.23 questions with the standard deviation of 30.42. 

We report the average and standard deviation of the accuracy achieved by the volunteers in Table~\ref{tab:human}. We notice the significant gap between the accuracies by the first and second groups, suggesting that the difficulty of question-answering grows as the length of the answer increases. Also, according to the F1 scores, we observe a large gap between the ASR and humans. This suggests the potential for the proposed SearchQA as a benchmark for advancing question-answering research. Overall, we found the performance of human volunteers much lower than expected and suspect the following underlying reasons. First, snippets are noisy, as they are often excerpts not full sentences. Second, human volunteers may have become exhausted over the trial. We leave more detailed analysis of the performance of human subjects on the proposed SearchQA for the future.


\subsection{Machine Baselines}

\paragraph{TF-IDF Max} 

An interesting property of the proposed SearchQA is that the context of each question-answer pair was retrieved by Google with the question as a query. This implies that the information about the question itself may be implicitly embedded in the snippets. We therefore test a naive strategy (TF-IDF Max) of selecting the word with the highest TF-IDF score in the context as an answer. Note that this can only be used for the questions with a unigram answer.

\paragraph{Attention Sum Reader}

Attention sum reader \citep[ASR,][]{kadlec2016text} is a variant of a pointer network \citep{vinyals2015pointer} that was specifically constructed to solve a cloze-style question-answering task. ASR consists of two encoding recurrent networks. The first network encodes a given context $c$, which is the concatenation of all the snippets in the case of SearchQA, into a set of hidden vectors $\{ h_j^c \}$, and the second network encodes a question $q$ into a single vector $h^q$. The dot product between each hidden vector from the context and the question vector is exponentiated to form word scores $\beta_j = \exp({h^q}^\top h_j^c)$. ASR then pulls these word scores by summing the scores of the same word, resulting in a set of unique word scores $\beta_i' = \sum_{j \in D_i} \beta_j$, where $D_i$ indicates where the word $i$ appears in the context. These unique-word scores are normalized, and we obtain an answer distribution $p(i|c, q) = \beta'_i/\sum_{i'} \beta'_{i'}$. The ASR is trained to maximize this (log-)probability of the correct answer word in the context. 

This vanilla ASR only works with a unigram answer and is not suitable for an $n$-gram answer. We avoid this issue by introducing another recurrent network which encodes the previous answer words $(\hat{a}_1, \ldots, \hat{a}_{l-1})$ into a vector $h^a$. This vector is added to the question vectors, i.e., $h^q\leftarrow h^q+h^a$. During training, we use the correct previou answer words, while we let the model, called n-gram ASR, predict one answer at a time until it predicts $\left<\text{answer}\right>$. This special token, appended to the context, indicates the end of the answer. 

We try both the vanilla and n-gram ASR's on the unigram-answer-only subset and on the whole set, respectively. We use recurrent networks with 100 gated recurrent units~\citep[GRU,][]{cho2014learning} for both unigram and $n$-gram models, respectively. We use Adam~\citep{kingma2014adam} and dropout~\citep{srivastava2014dropout} for training. 

\begin{table}[t]
\centering
\begin{tabular}{c c || c  c | c}
& & \multicolumn{2}{c|}{Unigram} & $n$-gram \\
Model & Set & Acc & Acc@5 & F1 \\
\hline\hline
TF-IDF & Valid & 13.0 & 49.3 & -- \\
Max & Test & 12.7 & 49.0 & -- \\
\hline
\multirow{2}{*}{ASR} & Valid & 43.9 & 67.3 & 24.2 \\
& Test & 41.3 & 65.1 & 22.8 
\end{tabular}
\caption{
\label{tab:asr}
The accuracies on the validation and test sets using the non-trainable baseline (TF-IDF Max) and the trainable baseline (ASR). We report top-1/5 accuracies for unigram answers, and otherwise, F1 scores.
}

\vskip -6mm

\end{table}

\paragraph{Result}

We report the results in Table~\ref{tab:asr}. We see that the attention sum reader is below human evaluation, albeit by a rather small margin. Also, TF-IDF Max scores are not on par when compared to ASR which is perhaps not surprising. Given the unstructured nature of SearchQA, we believe improvements on the benchmarks presented are crucial for developing a real-world Q\&A system.

\section{Conclusion}

We constructed a new dataset for question-answering research, called SearchQA. It was built using an in-production, commercial search engine. It closely reflects the full pipeline of a (hypothetical) general question-answering system, which consists of information retrieval and answer synthesis. We conducted human evaluation as well as machine evaluation. Using the latest technique, ASR, we show that there is a meaningful gap between humans and machines, which suggests the potential of SearchQA as a benchmark task for question-answering research. We release SearchQA publicly, including our own implementation of ASR and n-gram ASR in PyTorch.\footnote{http://pytorch.org/}

\section*{Acknowledgments}
KC thanks support by Google,  NVIDIA, eBay and Facebook. MD conducted this work as a part of DS-GA 1010: Independent Study in Data Science at the Center for Data Science, New York University.

\bibliography{emnlp2017}
\bibliographystyle{emnlp_natbib}

\end{document}